%% file: main.tex
\documentclass[10pt,twocolumn,letterpaper]{article}

\usepackage{iccv}              


\definecolor{iccvblue}{rgb}{0.21,0.49,0.74}
\usepackage[pagebackref,breaklinks,colorlinks,allcolors=iccvblue]{hyperref}

\usepackage{url}

\usepackage{algorithm}
\usepackage{algorithmic}
\usepackage{booktabs}
\usepackage{times}
\usepackage{epsfig}
\usepackage{amsmath}
\usepackage{amssymb}
\usepackage{multirow}
\usepackage[flushleft]{threeparttable}
\usepackage{pifont}

\usepackage{color, colortbl}
\usepackage{array}
\usepackage{makecell}
\usepackage[utf8]{inputenc} 
\usepackage[T1]{fontenc}

\definecolor{mygray}{gray}{.9}
\definecolor{mygreen}{rgb}{0, 0.6, 0}

\def\paperID{*****} 
\def\confName{ICCV}
\def\confYear{2025}

\title{
\includegraphics[height=15pt]{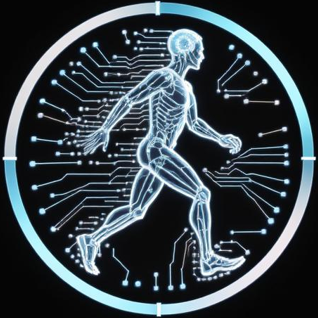} 
GaitCrafter: Diffusion Model for Biometric Preserving Gait Synthesis
}

\author{
Sirshapan Mitra \qquad Yogesh S. Rawat \\
CRCV, University of Central Florida \\
{\tt\normalsize \{sirshapan.mitra, yogesh\}@ucf.edu} \\
\normalsize\textbf{\url{https://sirsh07.github.io/research/gaitcrafter}} 
}

\begin{document}


\twocolumn[{%
\renewcommand\twocolumn[1][]{#1}%
\maketitle
\begin{center}
    \centering
    \captionsetup{type=figure}
    \includegraphics[width=1.0\linewidth]{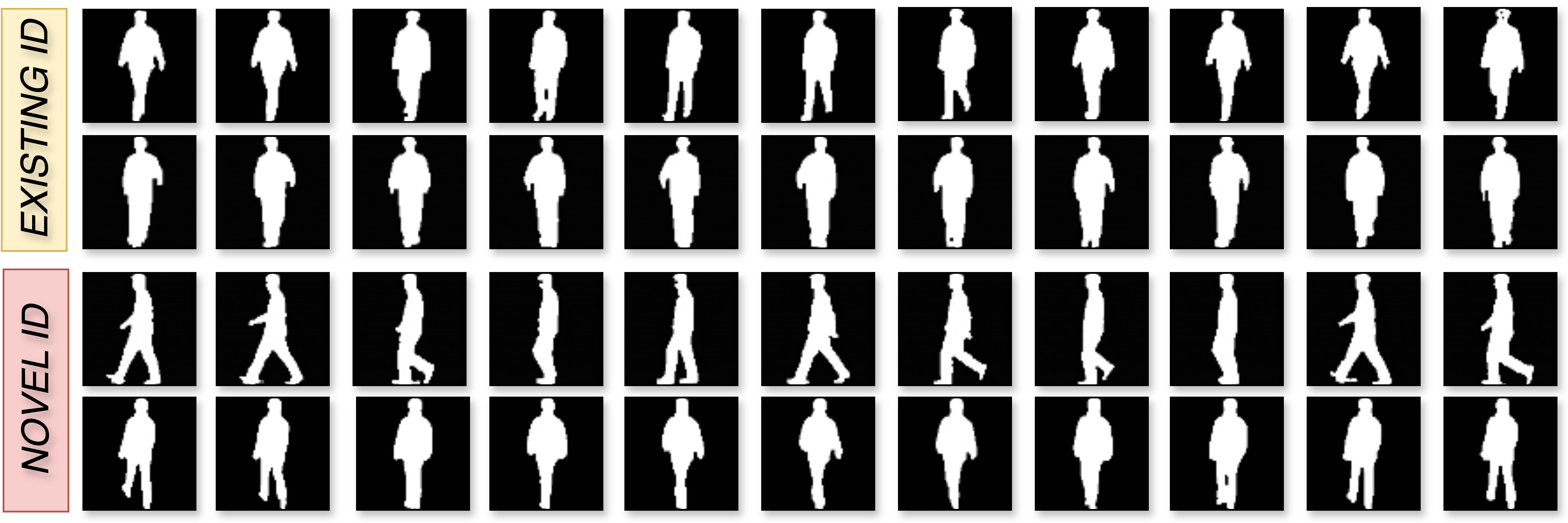} 
    \caption{ 
    \textbf{\textit{Generated gait sequences:}} GaitCrafter can generate both known and novel identities. Here we show generated 30-frame ($\sim$ 1 second) gait sequences for four different individuals, each covering a complete gait cycle. These sequences are temporally consistent and preserve both the structure and motion patterns characteristic of each subject's gait. As shown in the figure, the generated videos maintain the periodicity of walking and retain identity-specific biometric cues throughout the full sequence.
    }
    \label{fig:main_quant}
\end{center}%
}]

\input{sec/0_abstract}    
\input{sec/1_intro}

\input{sec/2_related_works}

\input{sec/3_method}
\input{sec/4_experiments}
{
    \small
    \bibliographystyle{ieeenat_fullname}
    \bibliography{main}
}

\end{document}

%% file: sec/0_abstract.tex
\begin{abstract}
Gait recognition is a valuable biometric task that enables the identification of individuals from a distance based on their walking patterns. However, it remains limited by the lack of large-scale labeled datasets and the difficulty of collecting diverse gait samples for each individual while preserving privacy. To address these challenges, we propose \textbf{GaitCrafter}, a diffusion-based framework for synthesizing realistic gait sequences in the silhouette domain. Unlike prior works that rely on simulated environments or alternative generative models, \textit{GaitCrafter} trains a video diffusion model from scratch, exclusively on gait silhouette data. Our approach enables the generation of temporally consistent and identity-preserving gait sequences. Moreover, the generation process is controllable-allowing conditioning on various covariates such as clothing, carried objects, and view angle. We show that incorporating synthetic samples generated by \textit{GaitCrafter} into the gait recognition pipeline leads to improved performance, especially under challenging conditions. Additionally, we introduce a mechanism to generate \textit{novel identities}-synthetic individuals not present in the original dataset-by interpolating identity embeddings. These novel identities exhibit unique, consistent gait patterns and are useful for training models while maintaining privacy of real subjects. Overall, our work takes an important step toward leveraging diffusion models for high-quality, controllable, and privacy-aware gait data generation. Github: \url{https://github.com/sirsh07/GaitCrafter}.



\end{abstract}



%% file: sec/1_intro.tex
\section{Introduction}
\label{sec:intro}






Gait recognition is the task of identifying individuals based on their body shape and walking patterns. As a biometric modality~\cite{nixon2006automatic}, it offers several advantages over traditional methods such as fingerprint or facial recognition\cite{yi2013towards, jain1997line} and is different from general video recognition tasks \cite{Kumar_2022_CVPR, modi2022video, Dave_2022_WACV, kumar2023benchmarking,  Singh_Rana_Kumar_Vyas_Rawat_2024, Kumar_2025_CVPR, kumar2025stable, kumar2025contextual, garg2025stpro}. Notably, gait can be captured unobtrusively from a distance, without requiring the subject’s cooperation or awareness, making it highly suitable for surveillance and non-intrusive biometric authentication.

Over the years, research in gait recognition has seen significant progress, evolving from handcrafted features and shallow classifiers to deep convolutional networks\cite{gaitgl} and, more recently, transformer-based architectures\cite{fan2023exploring}. Approaches in this domain vary widely, treating gait sequences either as unordered sets or temporally ordered sequences, and leveraging different modalities such as binary silhouettes, pose keypoints, or RGB frames. Among these, silhouette-based methods have emerged as a popular choice due to their compact representation, privacy-preserving nature, and strong performance across benchmarks \cite{gaitgl, Fan2020GaitPartTP, Chao2018GaitSetRG}

Despite recent advancements, gait recognition still faces several critical challenges. First, the collection and manual annotation of large-scale labeled datasets is both labor-intensive and expensive. While some datasets like GREW\cite{grew} and OUMVLP\cite{oumvlp} provide extensive labeled gait sequences, and others like GaitLU-1M\cite{gaitlu-1m} offer large-scale unlabeled data, the task inherently demands coverage of a wide range of covariates. These include variations in clothing, carried objects (e.g., bags), and viewing angles-factors that significantly impact recognition accuracy. This makes it difficult to scale purely through real-world data collection.


To address the limitations of limited labels and diverse sample collection in gait recognition, we explore diffusion-based models for generating high-quality synthetic gait sequences for both known and novel identities. Unlike prior works that rely on simulated environments~\cite{dou2021versatilegait} or alternative generative models such as VAEs~\cite{ma2023pedestrian}, our approach is the first to train a video diffusion model directly on silhouette data. This enables the generation of realistic, temporally consistent gait cycles in the silhouette domain. Beyond generation, we show that incorporating these synthetic sequences into the gait recognition pipeline improves downstream performance, particularly in low-label or privacy-constrained scenarios-demonstrating the effectiveness of synthetic data not just for visual quality but also for recognition utility.

We propose a diffusion-based framework for synthesizing novel gait sequences, with the goal of improving recognition performance-particularly in low-label regimes. Our main contributions are summarized as follows:
\begin{itemize}
    \item \textbf{Consistent gait sequence generation:} We propose GaitCrafter, a diffusion-based framework to generate realistic and temporally consistent gait sequences in the silhouette domain. Our model is trained from scratch and captures complete gait cycles for multiple human identities.
    
    \item \textbf{Controllable generation across covariates:} We enable fine-grained control over generated gait sequences by conditioning on covariates such as camera viewpoint, clothing, and carried objects. This allows us to simulate diverse walking conditions that are difficult to collect in real-world datasets.
    
    \item \textbf{Improving recognition with synthetic data:} We show that incorporating synthetic sequences into the training pipeline improves gait recognition performance, particularly under challenging conditions. Our method also supports the generation of novel identities, expanding the identity space and helping the model generalize better.
\end{itemize}


%% file: sec/2_related_works.tex
\section{Related Works}
\label{sec:relatedworks}

\begin{figure*}[ht]
    \centering
    \includegraphics[width=1.0\textwidth]{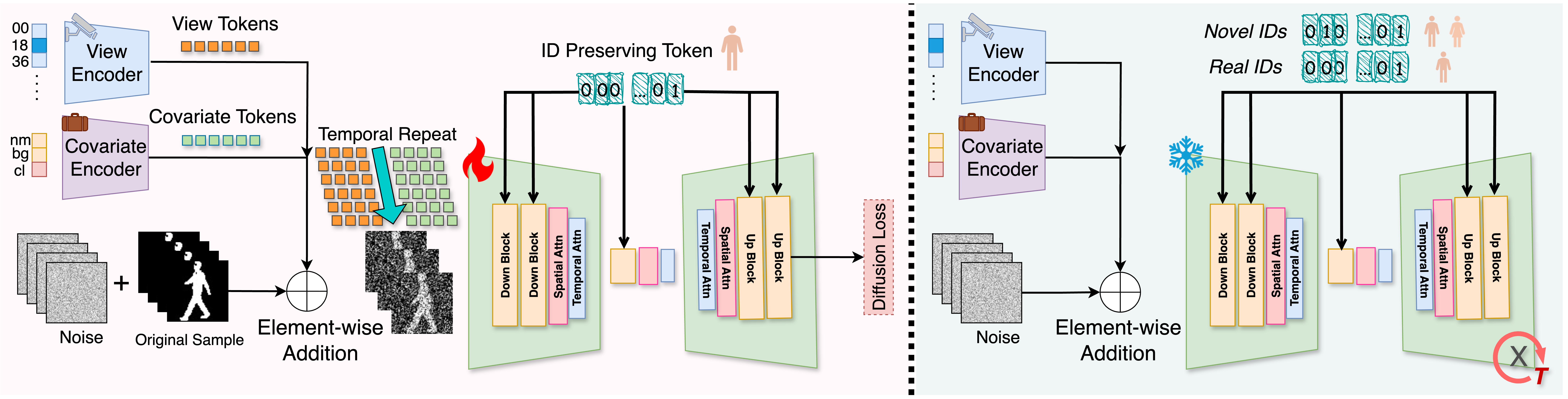} 
    \caption{
    \textbf{\textit{Overview of GaitCrafter:}} The left block shows the \textit{training stage}, where the model is trained using covariate, view, and identity conditions. At \textit{inference time} (right block), GaitCrafter can mix one-hot tokens of different IDs to generate a novel identity sample.
    }
    \label{fig:main_arch_subset}
\end{figure*}

\paragraph{Gait recognition} focuses on identifying individuals based on their walking patterns and body movements. This task can be approached through various modalities, including physical body structure and motion dynamics~\cite{chai2022lagrange}, binary silhouettes, or RGB video frames~\cite{Fan2020GaitPartTP, gaitgl, gaitbase, huang20213d, Wang2023DyGaitED, dou2022metagait}. Our work centers on the silhouette-based modality due to its balance between privacy and discriminative power.

Early methods addressed the task by capturing either global features~\cite{Chao2018GaitSetRG}, local discriminative parts~\cite{Fan2020GaitPartTP}, or a combination of both~\cite{gaitgl}. 

Inspired by image classification, GaitBase~\cite{gaitbase} employed a straightforward ResNet-based architecture~\cite{resnet} to learn effective gait embeddings. More recent advancements have\cite{dou2022metagait,Wang2023DyGaitED, Ma_2023_CVPR, Wang2023HierarchicalSR} emphasized dynamic feature adaptation. 

\noindent\textbf{Synthetic Data in Biometrics} 
Generative models have become an essential tool in biometric research, addressing challenges such as data scarcity, covariate variation, privacy preservation, and robust spoof‐detection.  Early work leveraged \textit{variational autoencoders} (VAEs) to synthesize low-resolution facial images, fingerprints, and irises for data augmentation~\cite{tu2019facial, li2018facial}.  Although VAEs provided controllable latent spaces, the visual fidelity and identity retention were limited. Moreover, previous works on biometrics have primarily focused on the RGB domain \cite{Pathak_2025_ICCV, Azad_2025_ICCV, Liang_2025_CVPR, Azad_2024_CVPR, pathak2025coarse}

\noindent\textbf{GAN-based Synthesis}
The introduction of \textit{generative adversarial networks} (GANs) elevated image realism.  StyleGAN-series models produced high-resolution faces that are now used for both augmentation and deepfake detection research~\cite{yang2023styleganex}.  In gait analysis, GaitGAN and its variants focused on silhouette translation across viewpoints or clothing conditions~\cite{yu2017gaitgan, yu2017gaitgan2}.  Despite their success, classical GANs often struggle with temporal consistency in video biometrics and require careful training to avoid mode collapse.

\noindent\textbf{Diffusion Models for Biometrics.}
Diffusion models have recently surpassed GANs in perceptual quality and likelihood estimation.  Face diffusion frameworks enable controllable editing (age, expression) while preserving identity embeddings~\cite{li2024id, shahreza2024hyperface}.  Video diffusion has been applied to talking-head generation, producing temporally coherent sequences for lip-reading and avatar animation~\cite{stypulkowski2024diffused}.  However, most diffusion work focuses on RGB data; dedicated silhouette‐based diffusion for gait, as proposed in our study, remains under-explored.

\noindent\textbf{Limitations and Open Problems.}
Key challenges persist across modalities: (i) \emph{identity preservation}-ensuring synthetic samples faithfully encode biometric signatures while remaining visually diverse; (ii) \emph{covariate control}-modulating pose, illumination, or accessories in a disentangled manner; and (iii) \emph{temporal coherence} for video biometrics such as gait and lip movement.  Our work addresses these gaps by training a silhouette-domain video diffusion model that (1) maintains identity features, (2) enables explicit covariate control (viewpoint, clothing, baggage), and (3) produces temporally consistent 30-frame gait cycles suitable for downstream recognition.

%% file: sec/3_method.tex
\section{Method}
\label{sec:method}



\begin{figure*}[t!]
    \centering
    \includegraphics[width=1.0\textwidth]{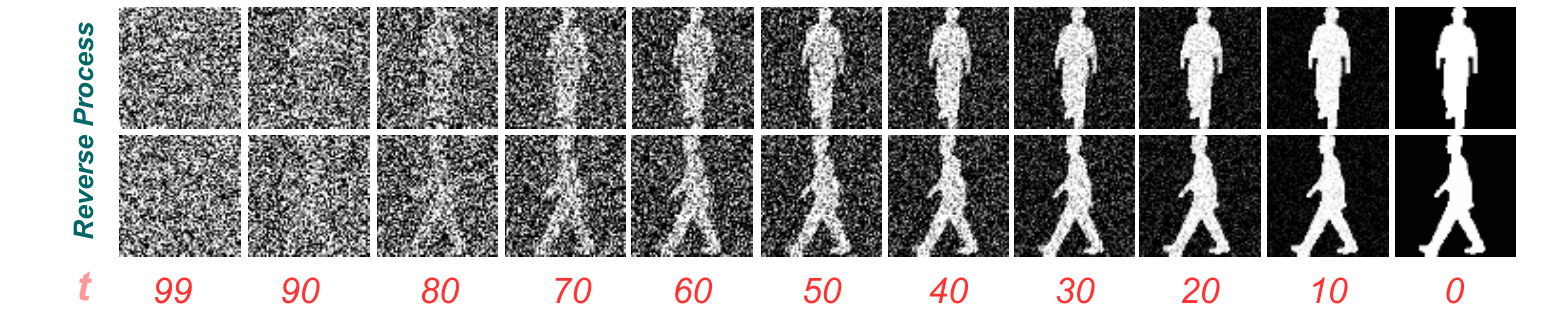} 
    \caption{
    \textbf{\textit{Overview of Reverse Process:}} We visualize the reverse diffusion process during generation. It is interesting to observe that starting from standard Gaussian noise, the model is able to recover a clean bimodal distribution characteristic of gait silhouettes-composed of sharp foreground and background regions. 
    }
    \label{fig:denoise}
    \vspace{-10pt}
\end{figure*}



\textbf{Overview:}
Gait recognition is fundamentally limited by the availability of labeled data-both in terms of the number of unique identities and the number of samples per identity. This scarcity makes it challenging to train robust models, especially under covariate variations such as clothing, view, and carried objects. To address this, we propose the use of synthetic data to augment training.
We leverage a diffusion model to generate realistic and temporally consistent gait sequences in the silhouette domain. Our hypothesis is that this synthetic data can improve gait recognition performance, particularly in low-label settings. We generate two types of synthetic data: (1) new samples for existing identities to increase intra-class variation, and (2) entirely new identities-referred to as novel IDs-to expand the identity space. Together, this synthetic data aim to improve gait models. 

\noindent \textbf{Background:} Diffusion model\cite{ddpm} involves two stages: forward process which does noise addition to the sample, and backward process, to denoise the samples in successive steps.  Given an initial data distribution $q$, the forward process incrementally introduces noise into data sampled from the distribution $y_0 \sim q(y_0)$. After adding noise at timestep $t$, $y_0$ becomes $y_t$. The backward process, or denoising, is designed to learn to remove this noise. Ultimately, during inference the denoising diffusion process generates a new data sample from a gaussian noise over a series of steps, effectively breaking down the complex task of modeling distributions into simpler, sequential denoising challenges.
This forward and backward process is described as,
\begin{equation}
  y_t = \sqrt{\alpha_t} y_{t-1} + \sqrt{1 - \alpha_t} \epsilon,
\end{equation}

\begin{equation}
    \hat{y}_{t-1} = \frac{1}{\sqrt{\alpha_t}} \left( y_t - \frac{1 - \alpha_t}{\sqrt{1 - \Bar{\alpha}_t}} \epsilon_\theta(y_t, t) \right),
\end{equation}
 where  $\epsilon \sim \mathcal{N}(0, I)$ and $\alpha_t$ are variance schedule coefficients, $\epsilon_\theta$ is learned noise prediction and $\Bar{\alpha}_t$ is the cumulative product of $\alpha_t$.


\subsection{GaitCrafter}
\label{subsec:GaitCrafter}





Video diffusion models typically operate either in pixel space\cite{Ho2022VideoDM} or in latent space\cite{zhou2022magicvideo,blattmann2023stable}. While latent-based models are generally faster and more resource-efficient, their limited capacity often makes it challenging to preserve fine-grained visual details, especially under complex conditions \cite{zhang2023show}. To better retain such details, we adopt a pixel-based video diffusion model \cite{Ho2022VideoDM} for synthetic gait sequence generation. This choice is particularly well-suited to gait recognition and biometric tasks, which require high fidelity in subtle motion and structural cues. Although pixel-based diffusion is computationally more demanding, we mitigate this overhead by employing the silhouette modality, which has a binary structure (mostly 0s and 1s) and a significantly simpler distribution compared to RGB inputs. This not only reduces the learning complexity but also accelerates the training process without sacrificing the preservation of discriminative features.


In our method (Figure~\ref{fig:main_arch_subset}), we build upon a 3D U-Net architecture within a video diffusion framework. The 3D U-Net with spatial and temporal attention layers enables our model to capture both spatial and temporal correlations across the entire gait sequence, which is essential for maintaining motion consistency and structural realism in generated videos.

We condition the generation process on three key factors that significantly impact gait appearance: (1) \textbf{person identity} (ID), (2) \textbf{view angle}, and (3) \textbf{covariates} such as clothing or carried objects. 

\noindent \textbf{ID Control} 
To represent identity, we use a one-hot encoded ID-preserving token ($x_{id}$), which is added as a conditioning signal to the U-Net blocks during diffusion. This design choice is motivated by the biometric nature of gait-each identity is expected to exhibit a unique and consistent motion pattern that the model should preserve.

\noindent \textbf{Condition Control} 
For view and covariate conditioning, we first encode the angle and covariate labels into separate latent tokens using a view encoder and a covariate encoder, respectively. These tokens are then temporally repeated to match the video length, resulting in 3D tensors: $x_{view}$ for view conditioning and $x_{covariate}$ for covariate conditioning. During generation, these tensors are added to the input-either the noise tensor during training or the video tensor during inference-to guide the model toward the desired attributes. This additive conditioning allows for fine-grained control over the output video’s viewpoint and covariate characteristics.

\begin{figure*}[ht]
    \centering
    \includegraphics[width=\linewidth]{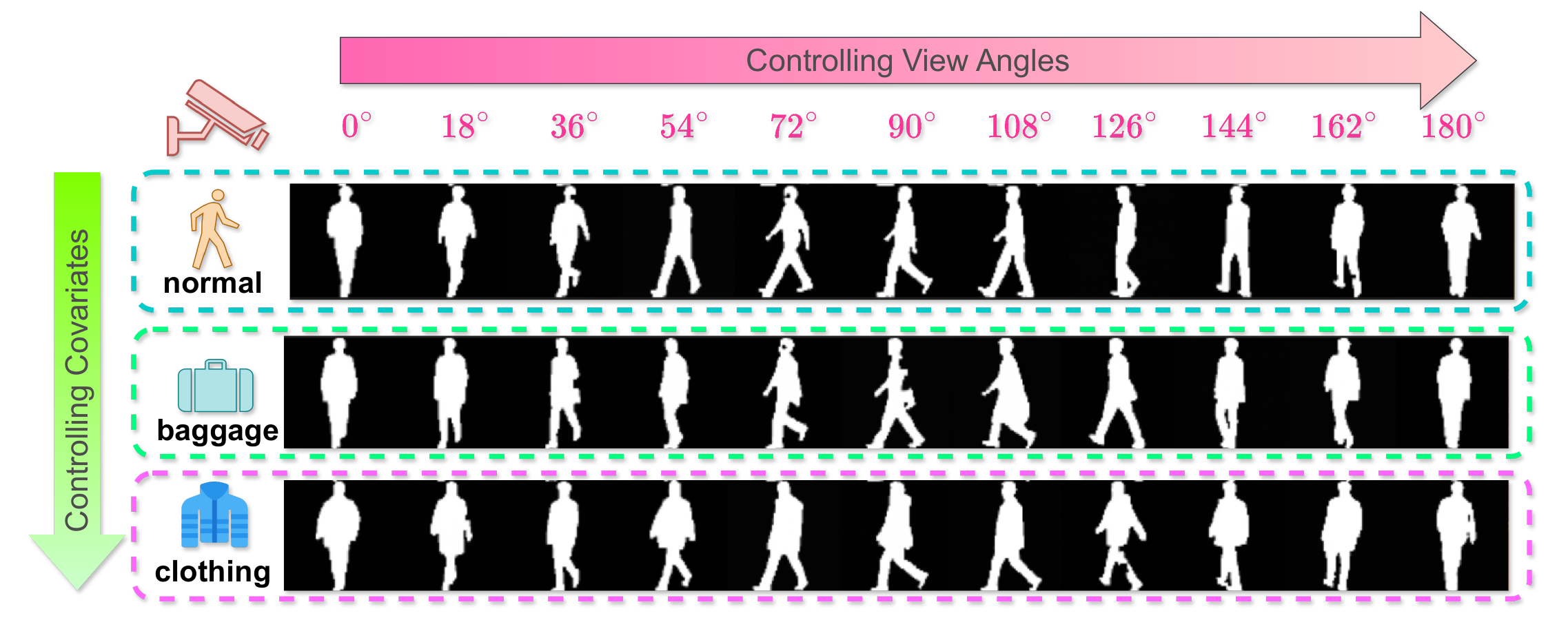} 
    \caption{
    \textbf{\textit{Diffusion Generated Sample:}}
    We show our diffusion-generated novel sample for different angles and clothing conditions. We show we can maintain consistent view angle and clothing condition. 
    }
    \label{fig:control_gait}
\end{figure*}

Overall, our conditioning strategy allows the diffusion model to synthesize gait sequences that are not only temporally coherent but also explicitly aligned with the desired identity, angle, and covariate conditions. This enables precise control during generation and improves the applicability of the synthetic sequences for training gait recognition models under varied scenarios.
\begin{equation}
    p_{\theta}(y_{t-1} | y_t, x_{cond}) = \mathcal{N}(y_{t-1}; \mu_{\theta}(x_{par}),   \Sigma_{\theta}(x_{par})),
\end{equation}
\begin{equation}
\begin{aligned}
    x_{par} = \{y_t, t,x_{id}, x_{cond}\}, 
    x_{cond} = \{x_{view}, x_{covariate}\}.
\end{aligned}
\end{equation}
Here, $p_\theta$ is the posterior with mean $\mu_{\theta}$ and variance $\Sigma_{\theta}$ modeled by the neural network with parameters $\theta$ and $y_t$ is the noisy sample at timestep $t$. 

We train the model by making it predict the noise using mean squared error loss, 
\begin{equation}
    L_{mse} = \mathbb{E}_{t \sim \mathcal{U}(1,T),y_0 \sim q(y_0),\epsilon} \left[ \left\| \epsilon - \theta(x_{par}) \right\|^2 \right]
\end{equation}
where, $\epsilon$ is the noise added, $x_{par}$ is the input to the diffusion model consisting of the noisy sample $y_{t}$, and timestep $t$. 

Finally, even though gait silhouettes exhibit a bimodal distribution-mostly consisting of foreground (value 1) and background (value 0)-we are still able to effectively model this distribution by simply adding Gaussian noise, following the standard diffusion process as shown in Fig~\ref{fig:denoise}. Despite the binary nature of the data, the model learns to capture the underlying structure and temporal dynamics of silhouette sequences, showing that the original diffusion formulation is sufficient for learning in this domain without requiring a specialized noise schedule or discretization strategy.


%% file: sec/4_experiments.tex
\section{Experiments and Results}
\label{sec:experiments}

\begin{table*}[t!]
\centering

  \small
  \begin{tabular}{ c| c| c c c c c c c c c c c | c}
    \toprule
    \multicolumn{2}{c|}{Methods}&  $0^{\circ}$ & $18^{\circ}$ & $36^{\circ}$ & $54^{\circ}$ & $72^{\circ}$ & $90^{\circ}$ & $108^{\circ}$ & $126^{\circ}$ & $144^{\circ}$ & $162^{\circ}$ & $180^{\circ}$ & Mean\\
    \midrule
   \multirow{3}{*}{NM} & $O$  & 95.1 & 98.3 & 98.9 & 97.9 & 96.2 & 94.1 & 97.2 & 98.9 & 99.4 & 98.8 & 94.1 & 97.2 \\
   & + $O_D$ & 95.1 & 97.5 & 98.6 & 97.4 & 95.6 & 93.0 & 96.0 & 98.4 & 98.9 & 97.8 & 92.2 & 96.4 \\
   & + $O_N$  & 94.5 & 97.7 & 98.3 & \underline{97.7} & \textbf{96.2} & \textbf{94.7} & \underline{97.1} & \textbf{98.9} & \underline{99.1} & \underline{98.5} & \underline{93.8} & \underline{97.1} \\
   \midrule
    \multirow{3}{*}{BG} & $O$  & 92.7 & 96.1 & 96.7 & 95.5 & 93.7 & 88.5 & 92.8 & 96.3 & 98.3 & 96.3 & 90.3 & 94.3 \\
   & + $O_D$ & 91.6 & 95.3 & 96.4 & 94.8 & 93.2 & 89.2 & 91.8 & \textbf{97.3} & \underline{98.2} & 96.1 & 89.8 & 93.9 \\
   & + $O_N$  & 91.4 & \underline{95.4} & \textbf{97.4} & \textbf{95.5} & \textbf{94.4} & \textbf{92.7} & \textbf{93.0} & 96.7 & 97.8 & \textbf{97.8} & \textbf{91.3} & \textbf{94.9} \\
   \midrule
    \multirow{3}{*}{CL} & $O$  & 76.3 & 89.3 & 91.0 & 89.3 & 84.4 & 77.4 & 83.3 & 87.9 & 87.6 & 84.1 & 69.1 & 83.6 \\
   & + $O_D$ & \textbf{77.1} & 90.4 & 92.5 & 89.5 & 85.4 & 79.6 & 82.7 & 87.1 & 90.7 & 83.8 & 70.9 & 84.5 \\
   & + $O_N$  & 75.8 & \textbf{90.6} & \textbf{93.3} & \textbf{90.6} & \textbf{87.0} & \textbf{80.8} & \textbf{85.7} & \textbf{88.7} & \textbf{91.6} & \textbf{86.2} & \textbf{71.0} & \textbf{85.6}  \\
    \bottomrule
  \end{tabular}
    \caption{
  \textit{\textbf{Impact of Synthetic data on gait recognition performance in CASIA-B}} with 100\% labeled data. $O$ means Original data, $O_D$ means diffusion generated original data and $O_N$ means diffusion generated novel data.}
    \label{tab:diffusion_setup_100per_lbldp}
    
\end{table*}

\begin{table*}[t]
\centering
  
  \small
  \begin{tabular}{ c| c| c c c c c c c c c c c | c}
    \toprule
    \multicolumn{2}{c|}{Methods}&  $0^{\circ}$ & $18^{\circ}$ & $36^{\circ}$ & $54^{\circ}$ & $72^{\circ}$ & $90^{\circ}$ & $108^{\circ}$ & $126^{\circ}$ & $144^{\circ}$ & $162^{\circ}$ & $180^{\circ}$ & Mean\\
    \midrule
   \multirow{3}{*}{NM} & $O$  & 82.3 & 92.0 & 94.3 & 91.4 & 84.8 & 82.9 & 85.6 & 92.3 & 94.4 & 92.0 & 75.8 & 87.9 \\
   & + $O_D$ & 82.6 & 88.9 & 94.0 & 90.5 & 85.4 & 82.9 & 86.4 & 92.1 & 93.6 & 90.0 & 77.8 & 87.7 \\
   & + $O_N$  & 76.4 & 88.2 & 93.3 & 90.8 & 84.3 & 78.6 & 83.0 & 92.5 & 94.7 & 88.9 & 75.6 & 86.0 \\
   \midrule
    \multirow{3}{*}{BG} & $O$  & 72.7 & 84.0 & 88.0 & 86.2 & 78.1 & 74.3 & 77.0 & 85.1 & 87.2 & 87.3 & 69.1 & 80.8 \\
   & + $O_D$ & 72.5 & 81.7 & 85.7 & 85.3 & 79.0 & 72.7 & 76.5 & 85.1 & 88.1 & 87.3 & 70.0 & 80.4 \\
   & + $O_N$  & 69.1 & 82.2 & 86.4 & 85.0 & 76.5 & 70.9 & 77.1 & 85.3 & 88.3 & 82.2 & 64.9 & 78.9 \\
   \midrule
    \multirow{3}{*}{CL} & $O$  & 49.3 & 67.1 & 68.9 & 65.0 & 59.1 & 52.1 & 53.3 & 59.6 & 62.0 & 58.8 & 44.5 & 58.2 \\
   & + $O_D$ & 48.6 & 62.1 & 66.5 & 63.3 & 57.6 & 55.4 & 55.6 & 60.4 & 64.6 & 58.2 & 42.1 & 57.8 \\
   & + $O_N$  & 49.2 & 64.2 & 67.4 & 68.5 & 63.7 & 59.5 & 60.7 & 67.0 & 67.5 & 59.9 & 43.8 & 61.0 \\
    \bottomrule
  \end{tabular}

\caption{
  \textit{\textbf{Impact of Synthetic data Open-set settings on CASIA-B}} with 20\% labeled subset. $O$ means Original CASIA-B, $O_D$ means diffusion generated original data and $O_N$ means diffusion generated novel data.}
  \label{tab:diffusion_setup_20per_lblpid}
\end{table*}

\noindent \textbf{Diffusion setup: }We employ a 3D U-Net architecture for our diffusion model, with an input dimension of 64 and channel multipliers set to $(1, 2, 4, 8)$. The model is conditioned on identity information using one-hot encodings, where the conditional dimension corresponds to the number of person IDs (PIDs) in the training data-74 for the full (100\%) setup and 15 for the 20\% setup. Each generated video consists of 30 frames. We train the model using the Adam optimizer with a learning rate of $1 \times 10^{-4}$, a batch size of 4, and 100 diffusion time steps.

The diffusion model is trained for 100K iterations with identity conditioning. Once trained, the model is used to generate synthetic samples for both original IDs $(O_D)$ and novel IDs $(N_D)$. The set $(O_D)$ includes the same number of IDs as in the original training data, while $(N_D)$ includes one fewer (i.e., $(|N_D| = |O_D| - 1)$). This results in an augmented dataset of size $(2 \times \text{PIDs} - 1)$.

\noindent \textbf{Experiment setup: } 
For all our analysis, we use the CASIA-B dataset\cite{casia} and the GaitGL\cite{gaitgl} model. GaitGL is one of the best-performing models on CASIA-B, and the dataset itself is relatively limited in both scale and diversity-making it a suitable benchmark for evaluating the effectiveness of synthetic data. 

We investigate two additional setups of incorporating synthetic data into gait recognition:

\begin{itemize}
    \item \textbf{Open Set:} We simulate scenarios with limited identity diversity (e.g., 20\% of total PIDs) and assess whether synthetic data can compensate for the restricted number of IDs.
    \item \textbf{Closed Set:} We examine performance under limited sample availability per identity (e.g., 20\% of samples per ID), testing whether synthetic augmentation improves model robustness in this low-sample regime.
\end{itemize}

It is important to note that, in both the open set and closed set cases, the diffusion model is trained using only the limited data, and the gait recognition model is trained on a combination of the limited real data and the corresponding synthetic data.


\subsection{Results and Analysis}


First, in Section~\ref{sec:qualitative}, we present the generative and visual results of our model, highlighting the quality and consistency of the synthesized gait sequences. Next, in Section~\ref{sec:quantitative}, we provide a quantitative evaluation of the generated samples. In Section~\ref{sec:downstream}, we analyze the impact of incorporating synthetic data into downstream gait recognition models. Finally, in Section~\ref{sec:discussion}, we discuss key challenges and limitations encountered during our study.

\subsubsection{Qualitative Analysis of synthetic data: } 
\label{sec:qualitative}

\textbf{Consistent Gait Generation.} Most existing open-source video diffusion models~\cite{blattmann2023stable, chen2024videocrafter2} struggle to generate videos with more than 25 frames. Moreover, these models are generally trained on RGB datasets and are not well-suited for producing temporally consistent silhouette sequences, which are crucial for gait recognition tasks. In contrast, our proposed gait diffusion model is specifically trained to generate 30-frame silhouette sequences, corresponding to a full gait cycle~\cite{Chao2018GaitSetRG}. This design choice ensures that each generated sequence captures the complete periodic motion of a walking individual. As illustrated in Figure~\ref{fig:main_quant}, our model successfully generates consistent 30-frame gait cycles across multiple identities, demonstrating both temporal coherence and structural stability in the silhouette domain.

\begin{figure*}[t!]
    \centering
    \begin{minipage}{0.32\textwidth}
        \centering
        \includegraphics[width=1.0\textwidth]{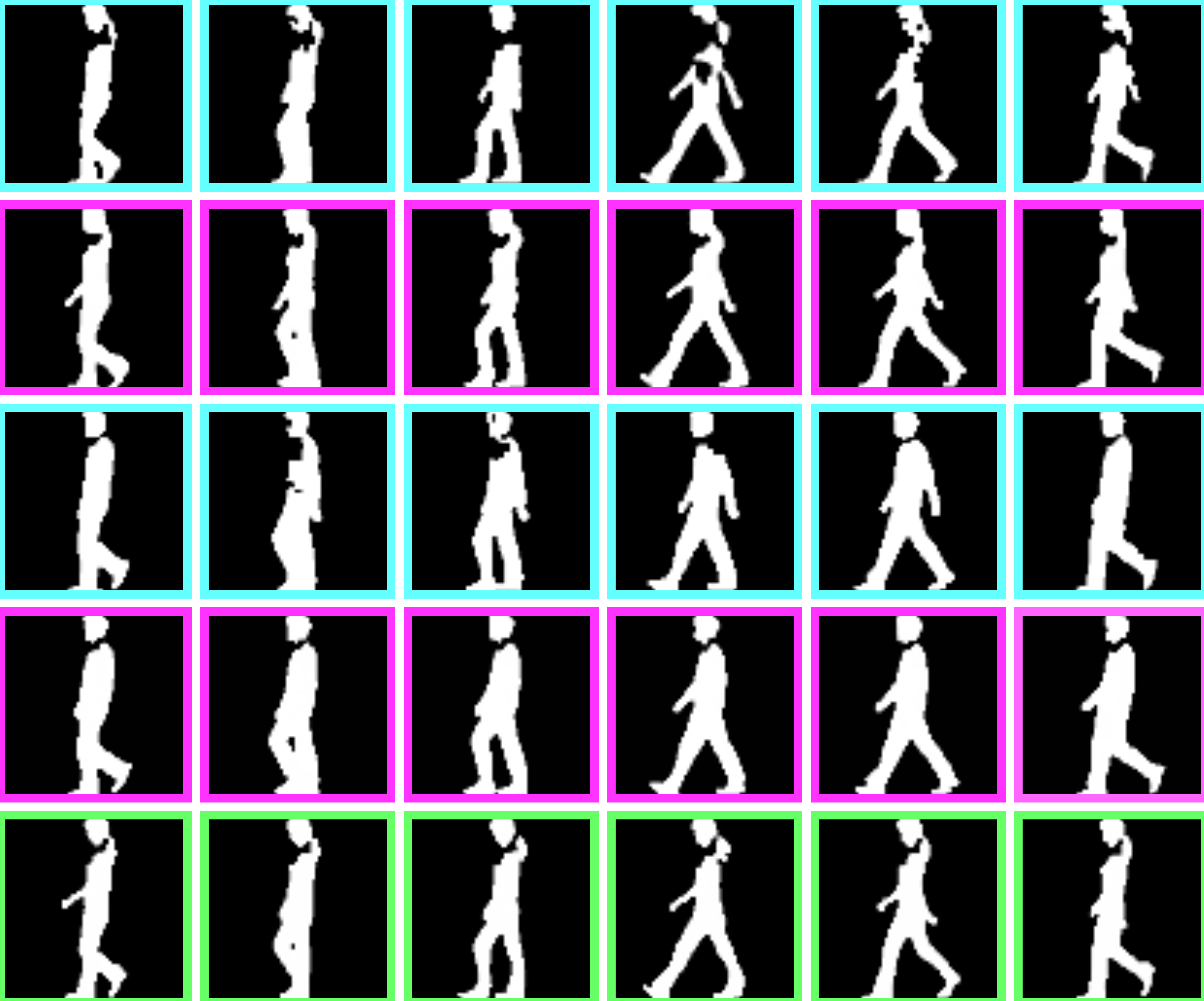}
    \end{minipage}
    \hfill
    \begin{minipage}{0.32\textwidth}
        \centering
        \includegraphics[width=1.0\textwidth]{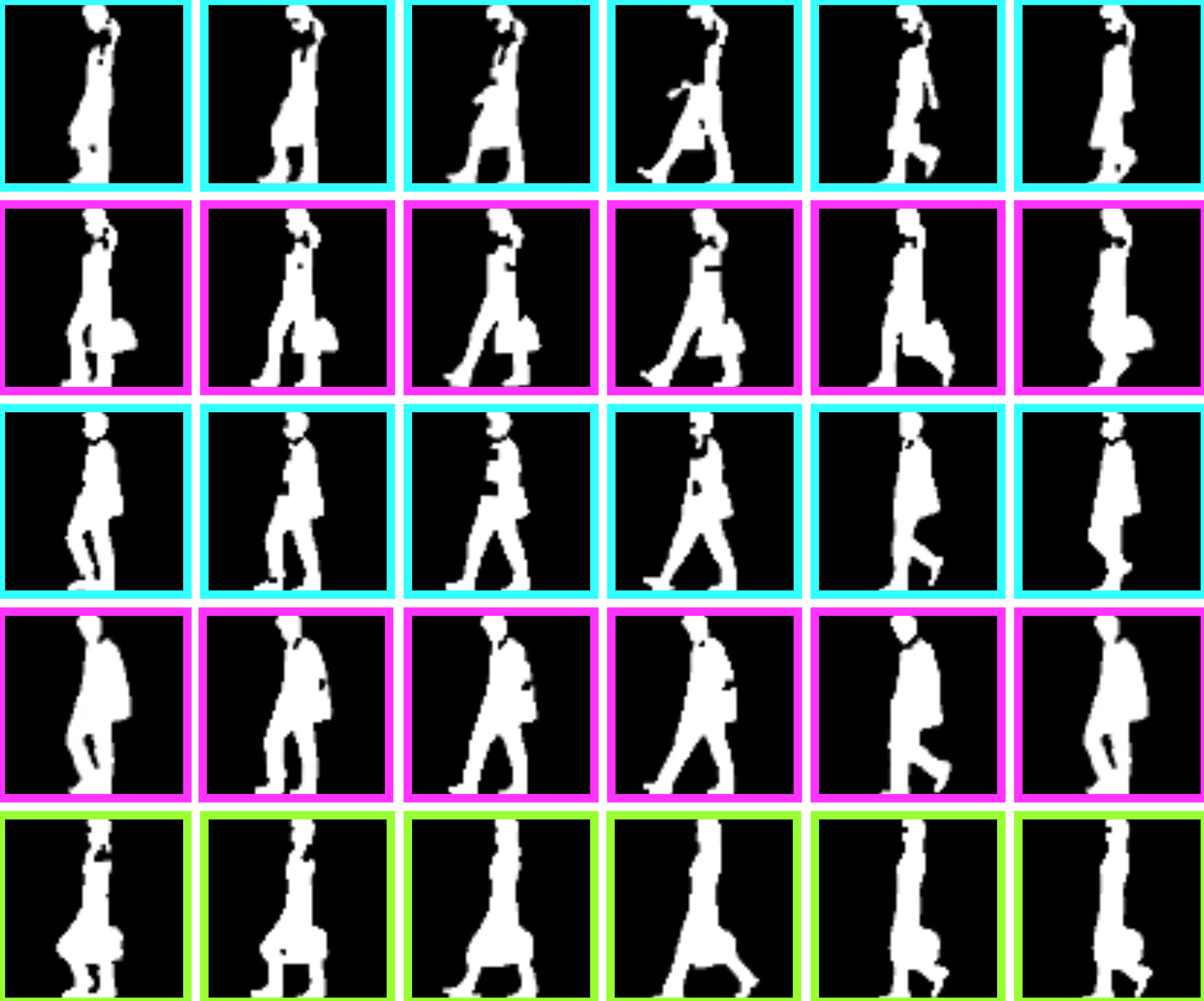}
    \end{minipage}
    \hfill
    \begin{minipage}{0.32\textwidth}
        \centering
        \includegraphics[width=1.0\textwidth]{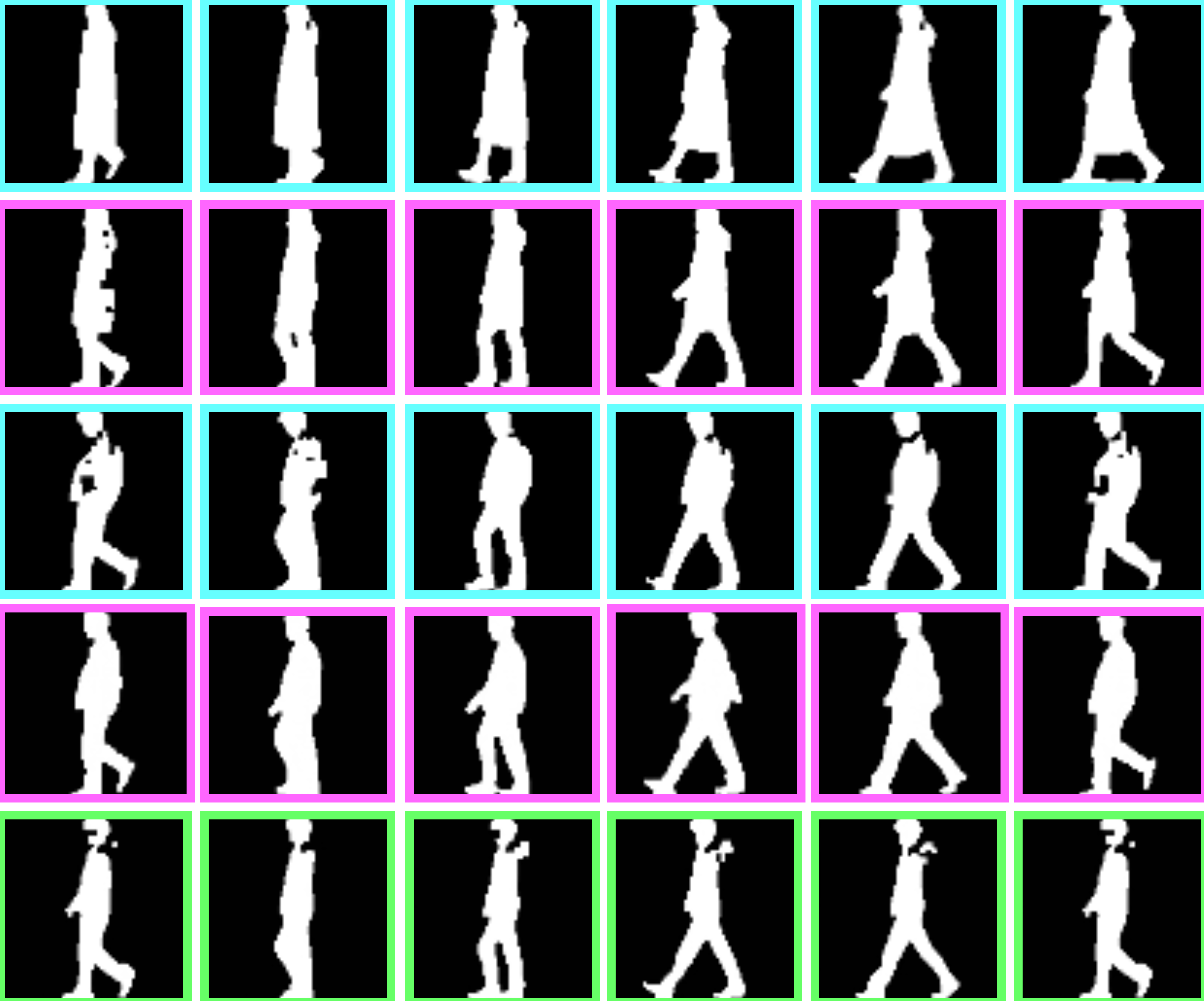}
    \end{minipage}
    \caption{
    \textbf{\textit{Discussions:}}
    \textit{Synthetic gait sequences:} Rows 1 and 3 in \textcolor[HTML]{66FFFF}{blue} show original samples, and rows 2 and 4 in 
\textcolor[HTML]{FF66FF}{pink} show corresponding generated samples for the same IDs. The last row shows a generated novel ID sequence in \textcolor[HTML]{66FF66}{green}. Left: NM condition. Middle: BG condition. Right: CL condition.
    }
    \label{fig:fig_discussion}
\end{figure*}

\noindent\textbf{Controllable Gait Generation.} Gait recognition is heavily influenced by covariates such as clothing, baggage, and camera angle. These factors introduce large intra-class variations, and handling them is essential for building robust gait recognition models. However, in real-world datasets, it's often difficult to collect gait sequences of the same individual under multiple covariate conditions. 

In Figure~\ref{fig:control_gait}, we show that our method can generate gait sequences while controlling for key covariates. Specifically, we demonstrate the ability to modify view angle, clothing, and baggage conditions for the same identity. This controllability is important because it allows us to generate missing variations for an individual, even if those conditions were not present in the original data. Our model learns to generalize across conditions and synthesize diverse yet identity-consistent sequences.

\noindent\textbf{Preserving Gait Biometric} 
A fundamental question when generating gait sequences is whether the identity-specific gait biometric is preserved. To validate this, we compare the embeddings of real and synthetic sequences using a pretrained gait recognition model. The idea is simple: if the synthetic samples truly preserve the identity, their embeddings should be close to the real samples of the same individual in the feature space.

\begin{figure}[t!]
    \centering
    \includegraphics[width=1.0\linewidth]{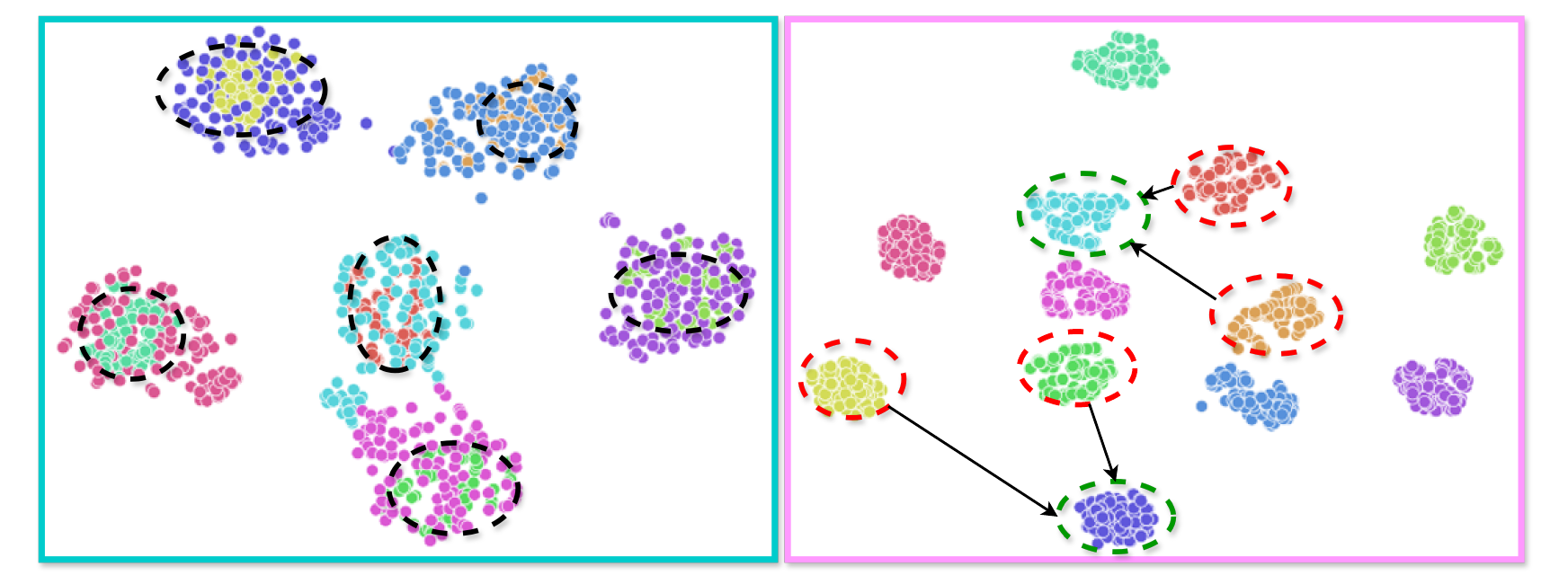} 
    \caption{ 
    \textbf{\textit{tSNE visualization of synthetic samples:}}
    The left plot shows real sample for known IDs, circled in black, demonstrating \textit{biometrics} is preserved for known IDs. In the right plot, we show known IDs circled in \textcolor{red}{red}, and generated novel IDs using those original ones in \textcolor{green}{green}.
    }
    \label{fig:tsne_diff_od}
\end{figure}

We use a GaitGL model pretrained on real CASIA-B data to extract gait embeddings. Since this model is trained to cluster similar gait patterns together, it serves as a good measure of biometric consistency. In Figure~\ref{fig:tsne_diff_od}, we show a t-SNE plot of the embeddings from both real and synthetic data. We observe that the synthetic samples cluster tightly around their corresponding real samples, indicating strong identity preservation.

This result supports our hypothesis that the diffusion-based synthetic sequences not only look realistic but also maintain the underlying gait signature required for recognition.

\noindent\textbf{Generating Novel Identities.} 
During training, we use one-hot embeddings to represent individual IDs in the diffusion model. A natural question that arises is-what happens if we activate multiple IDs at once during inference? We experiment by setting multiple entries in the one-hot embedding to 1, effectively blending identity codes. Interestingly, we observe that this results in the generation of new, distinct gait sequences.

As shown in Figure~\ref{fig:tsne_diff_od}, these mixed-ID samples form separate clusters in the embedding space, distinct from the original identities. This suggests that the model is not merely averaging existing identities, but learning to synthesize new, coherent gait patterns. We refer to these as \textit{novel IDs}.

This has important implications for gait recognition, which inherently suffers from a limited number of labeled identities. By generating novel IDs that are consistent, identity-distinct, and controllable under different covariates, we can significantly expand the training distribution. In Figure~\ref{fig:fig_discussion}, we show generated samples for two known IDs under baggage and clothing variations. In the last row, we present the novel ID generated by combining the two identity embeddings, and find that it results in a third, visually coherent and structurally consistent gait cycle. This opens up a promising direction for synthetic identity augmentation in biometric tasks.



\subsubsection{Quantitative Analysis: }
\label{sec:quantitative}

Since it is difficult to quantify the similarity of silhouette sequences using models trained on RGB images, we use a variant of the CLIP score tailored for gait. Specifically, we measure the cosine similarity between embeddings extracted from a pretrained gait recognition model using real samples for both real and generated samples. We refer to this metric as \textit{Gait Biometric Similarity (GBS)}, which captures how well the synthetic samples preserve identity-specific gait features. In our experiments, we use GBS to compare the quality of samples generated by different diffusion models and report the results as a measure of biometric consistency. 

For our study, we explore three different conditioning setups for identity representation within the diffusion model: \textit{Baseline (Condition Concatenation):} We follow a standard conditional diffusion setup where all conditioning variables-including identity, view, and covariate-are concatenated into a single input token. \textit{Gait Token as Identity:} We use pretrained gait tokens as identity tokens. \textit{MLP-Based ID Encoder:} We introduce a lightweight MLP to encode one-hot identity vectors into a learnable identity token space. 


These setups help us study how different forms of identity encoding impact the quality, consistency, and controllability of the generated gait sequences. 


\begin{table}

    \centering
    \renewcommand{\arraystretch}{1.1}
    \scalebox{0.97}{
    
\begin{tabular}{l|cccc}
\specialrule{1.5pt}{0pt}{0pt}

\rowcolor{mygray}Model & Baseline & Gait Tokens & MLP & Ours \\
\hline
GBS Score   & 0.674    & 0.543       & 0.222 & 0.753 \\
\specialrule{1.5pt}{0pt}{0pt}
\end{tabular}}
\caption{GBS Score comparison for different diffusion setups.}
\label{tab:gbs}
\vspace{-15pt}
\end{table}

\subsubsection{Effect of Downstream Tasks: }
\label{sec:downstream}

\noindent\textbf{Impact of Increasing Synthetic Samples per ID.}
Our first analysis investigates the effect of increasing the number of samples for each identity. In Table~\ref{tab:diffusion_setup_100per_lbldp}, we report the performance of a gait recognition model trained on both real data and diffusion-generated sequences corresponding to the same (real) IDs. We observe a modest improvement of 0.9\% under the clothing (CL) condition-one of the most challenging scenarios in the CASIA-B dataset. However, CASIA-B already contains multiple samples per subject, so simply adding more sequences for existing IDs does not significantly improve performance across all conditions.

\noindent\textbf{Impact of Increasing Number of Synthetic IDs.}
Next, we explore the effect of increasing the number of unique identities by introducing synthetic IDs-referred to earlier as \textit{novel IDs}. We generate multiple samples for these new identities, which remain consistent and preserve key gait characteristics. In Table~\ref{tab:diffusion_setup_100per_lbldp}, we show that training the model with a mix of real and novel IDs leads to a larger gain: a 2\% improvement under the clothing condition and a 0.9\% boost in overall average accuracy, using the same model architecture. These results highlight the benefit of synthetic identity expansion as a more effective strategy than simply increasing sample count for existing individuals.

\begin{figure*}[t!]
    \centering
    \includegraphics[width=\linewidth]{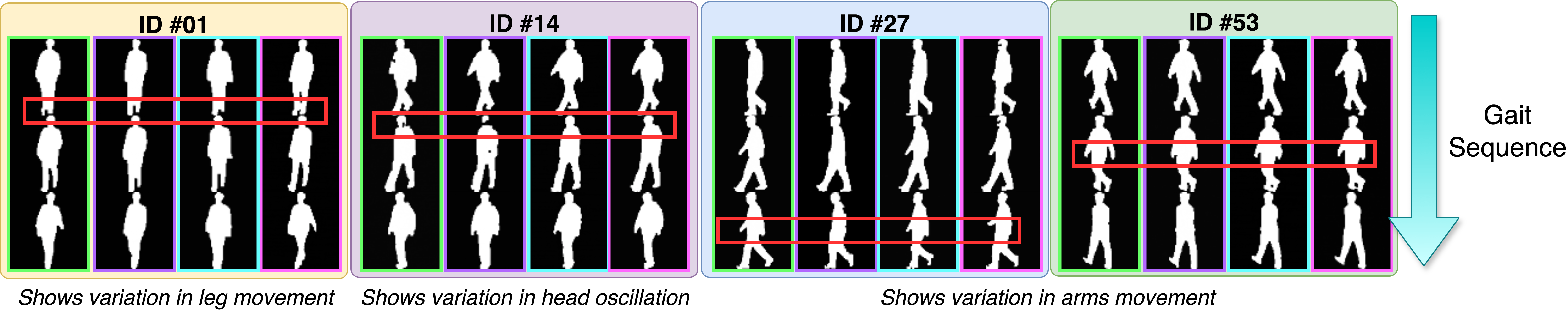} 
    \caption{
    \textbf{\textit{Different Generation using Different Seeds:}}
    Different Columns in the same ID, shows generation using different random seeds. We show different seeds generates small variation in gait pattern.
    }
    \label{fig:diff_seed}
\end{figure*}


\noindent\textbf{Closed Set vs. Open Set: Understanding the Role of Synthetic Data.}
One of the primary motivations for generating synthetic data in gait recognition is to support learning under limited-label scenarios. Synthetic data can be helpful in two major conditions: (1) when there are few samples available per individual, and (2) when the total number of identities (IDs) is limited. We refer to these two setups as the \textit{closed-set} and \textit{open-set} cases, respectively.

In the closed-set setup, we train the diffusion model on all identities but restrict the number of training samples to 20\% per ID. We then generate additional synthetic samples for these same identities to increase the per-ID data volume. We observed that adding synthetic samples per identity in this setting does not lead to any significant improvement in performance. In contrast, we explore the open-set setup by training the diffusion model on only a small subset of identities-20\% of the total, which corresponds to around 15 IDs. We then generate synthetic data for novel IDs not seen during training. In Table~\ref{tab:diffusion_setup_20per_lblpid}, we show that adding these synthetic novel IDs improves performance, especially under challenging covariates such as the clothing (CL) condition.


\subsubsection{Discussions: }
\label{sec:discussion}

\begin{table}
    \centering
    \renewcommand{\arraystretch}{1.1}
    \scalebox{0.97}{
    
        \begin{tabular}{c|ccc|c}
\specialrule{1.5pt}{0pt}{0pt}
\rowcolor{mygray}  Data & \multicolumn{3}{c|} {Score}  & avg \# of Frame  \\
\rowcolor{mygray} & NM & BG & CL  &   \\
\hline
Original ($O$) & 97.2 & 94.3 & 83.6 & 90 \\
Novdel Ids ($O_N$) & 93.4 & 89.5 & 75.6 & 30 \\

\specialrule{1.5pt}{0pt}{0pt}
\end{tabular}}
\caption{ \textbf{\textit{Effect of Only Novel Id's:}} We report the performance of a model trained only on synthetic novel IDs and evaluated on real test data. This setup helps us understand whether synthetic identities alone are sufficient for learning meaningful gait features.}

 	\label{tab:novtab}
\vspace{-15pt}
\end{table}

\textbf{Impact of Training Only on Synthetic Data.}
We further evaluate the effectiveness of our synthetic data by training the gait recognition model using only the synthetic novel IDs, without any real data. The goal is to see whether the synthetic samples alone are sufficient to preserve gait biometric patterns and enable downstream learning. This setup is particularly useful in scenarios where access to real biometric data is restricted due to privacy or legal concerns-making synthetic data a viable alternative.

Although the model is trained purely on synthetic data, we evaluate its performance on real test samples. As shown in Table~\ref{tab:novtab}, the model achieves performance comparable to training with real data. This demonstrates that the synthetic sequences not only retain discriminative gait features but also align closely with the real data distribution, allowing the model to generalize across domains without significant degradation. This small 6\% gap can be attributed to the fact that real CASIA-B gait videos contain more than 90 frames, offering higher motion diversity using temporal crops, whereas our synthetic sequences are limited to 30 frames. Despite this, the synthetic data still captures the essential characteristics of gait and proves effective for training in low-label or privacy-constrained settings.

\noindent\textbf{Multi-Sample Generation Using Different Seeds.}
In real gait recognition datasets, each video often contains more than 30 frames. This allows for temporal cropping during training, which introduces natural variation and helps improve model robustness. However, our diffusion model is designed to generate exactly 30-frame sequences-equivalent to a single gait cycle-so temporal cropping is not applicable to synthetic data.

To introduce variation and enrich the training set, we generate multiple versions of each sample by varying the random seed during inference as shown in Fig~\ref{fig:diff_seed}. Each seed produces a slightly different version of the walking sequence, even under the same identity and covariate conditions. We generate 5 such variations per setting. These samples maintain overall gait structure but introduce subtle changes in motion dynamics.

We find that training the gait recognition model with these additional variations improves performance compared to using just a single generated sequence per identity. This suggests that even small perturbations introduced via sampling help the model generalize better, compensating for the lack of temporal augmentation in fixed-length synthetic sequences.


\section{Conclusion}
We introduce GaitCrafter, a diffusion-based framework for generating realistic and controllable gait sequences in the silhouette domain. By training a video diffusion model from scratch, GaitCrafter enables identity-aware synthesis with covariate control, addressing key challenges in data scarcity and diversity. Our experiments demonstrate that incorporating synthetic data from GaitCrafter improves gait recognition performance, especially in low-label and semi-supervised settings. This work highlights the potential of diffusion models in advancing gait recognition through scalable and high-quality data generation.

%% file: main.bbl
\begin{thebibliography}{49}
\providecommand{\natexlab}[1]{#1}
\providecommand{\url}[1]{\texttt{#1}}
\expandafter\ifx\csname urlstyle\endcsname\relax
  \providecommand{\doi}[1]{doi: #1}\else
  \providecommand{\doi}{doi: \begingroup \urlstyle{rm}\Url}\fi

\bibitem[Azad and Rawat(2024)]{Azad_2024_CVPR}
Shehreen Azad and Yogesh~Singh Rawat.
\newblock Activity-biometrics: Person identification from daily activities.
\newblock In \emph{Proceedings of the IEEE/CVF Conference on Computer Vision and Pattern Recognition (CVPR)}, pages 287--296, 2024.

\bibitem[Azad and Rawat(2025)]{Azad_2025_ICCV}
Shehreen Azad and Yogesh~S Rawat.
\newblock Disenq: Disentangling q-former for activity-biometrics.
\newblock In \emph{Proceedings of the IEEE/CVF International Conference on Computer Vision (ICCV)}, 2025.

\bibitem[Blattmann et~al.(2023)Blattmann, Dockhorn, Kulal, Mendelevitch, Kilian, Lorenz, Levi, English, Voleti, Letts, et~al.]{blattmann2023stable}
Andreas Blattmann, Tim Dockhorn, Sumith Kulal, Daniel Mendelevitch, Maciej Kilian, Dominik Lorenz, Yam Levi, Zion English, Vikram Voleti, Adam Letts, et~al.
\newblock Stable video diffusion: Scaling latent video diffusion models to large datasets.
\newblock \emph{arXiv preprint arXiv:2311.15127}, 2023.

\bibitem[Chai et~al.(2022)Chai, Li, Zhang, Li, and Wang]{chai2022lagrange}
Tianrui Chai, Annan Li, Shaoxiong Zhang, Zilong Li, and Yunhong Wang.
\newblock Lagrange motion analysis and view embeddings for improved gait recognition.
\newblock In \emph{2022 IEEE/CVF Conference on Computer Vision and Pattern Recognition (CVPR)}, pages 20217--20226, 2022.

\bibitem[Chao et~al.(2018)Chao, He, Zhang, and Feng]{Chao2018GaitSetRG}
Hanqing Chao, Yiwei He, Junping Zhang, and Jianfeng Feng.
\newblock Gaitset: Regarding gait as a set for cross-view gait recognition.
\newblock In \emph{AAAI Conference on Artificial Intelligence}, 2018.

\bibitem[Chen et~al.(2024)Chen, Zhang, Cun, Xia, Wang, Weng, and Shan]{chen2024videocrafter2}
Haoxin Chen, Yong Zhang, Xiaodong Cun, Menghan Xia, Xintao Wang, Chao Weng, and Ying Shan.
\newblock Videocrafter2: Overcoming data limitations for high-quality video diffusion models.
\newblock In \emph{Proceedings of the IEEE/CVF Conference on Computer Vision and Pattern Recognition}, pages 7310--7320, 2024.

\bibitem[Dave et~al.(2022)Dave, Scheffer, Kumar, Shiraz, Rawat, and Shah]{Dave_2022_WACV}
Ishan Dave, Zacchaeus Scheffer, Akash Kumar, Sarah Shiraz, Yogesh~Singh Rawat, and Mubarak Shah.
\newblock Gabriellav2: Towards better generalization in surveillance videos for action detection.
\newblock In \emph{Proceedings of the IEEE/CVF Winter Conference on Applications of Computer Vision (WACV) Workshops}, pages 122--132, 2022.

\bibitem[Dou et~al.(2021)Dou, Zhang, Zhang, Zhao, Li, Qin, Wu, Dong, and Li]{dou2021versatilegait}
Huanzhang Dou, Wenhu Zhang, Pengyi Zhang, Yuhan Zhao, Songyuan Li, Zequn Qin, Fei Wu, Lin Dong, and Xi Li.
\newblock Versatilegait: a large-scale synthetic gait dataset with fine-grainedattributes and complicated scenarios.
\newblock \emph{arXiv preprint arXiv:2101.01394}, 2021.

\bibitem[Dou et~al.(2023)Dou, Zhang, Su, Yu, and Li]{dou2022metagait}
Huanzhang Dou, Pengyi Zhang, Wei Su, Yunlong Yu, and Xi Li.
\newblock Metagait: Learning to learn an omni sample adaptive representation for gait recognition.
\newblock \emph{ArXiv}, abs/2306.03445, 2023.

\bibitem[Fan et~al.(2020)Fan, Peng, Cao, Liu, Hou, Chi, Huang, Li, and He]{Fan2020GaitPartTP}
Chao Fan, Yunjie Peng, Chunshui Cao, Xu Liu, Saihui Hou, Jiannan Chi, Yongzhen Huang, Qing Li, and Zhiqiang He.
\newblock Gaitpart: Temporal part-based model for gait recognition.
\newblock \emph{2020 IEEE/CVF Conference on Computer Vision and Pattern Recognition (CVPR)}, pages 14213--14221, 2020.

\bibitem[Fan et~al.(2023{\natexlab{a}})Fan, Hou, Huang, and Yu]{fan2023exploring}
Chao Fan, Saihui Hou, Yongzhen Huang, and Shiqi Yu.
\newblock Exploring deep models for practical gait recognition.
\newblock \emph{arXiv preprint arXiv:2303.03301}, 2023{\natexlab{a}}.

\bibitem[Fan et~al.(2023{\natexlab{b}})Fan, Hou, Wang, Huang, and Yu]{gaitlu-1m}
Chao Fan, Saihui Hou, Jilong Wang, Yongzhen Huang, and Shiqi Yu.
\newblock Learning gait representation from massive unlabelled walking videos: A benchmark.
\newblock \emph{IEEE Transactions on Pattern Analysis and Machine Intelligence}, 45\penalty0 (12):\penalty0 14920--14937, 2023{\natexlab{b}}.

\bibitem[Fan et~al.(2023{\natexlab{c}})Fan, Liang, Shen, Hou, Huang, and Yu]{gaitbase}
Chao Fan, Junhao Liang, Chuanfu Shen, Saihui Hou, Yongzhen Huang, and Shiqi Yu.
\newblock Opengait: Revisiting gait recognition towards better practicality.
\newblock In \emph{Proceedings of the IEEE/CVF Conference on Computer Vision and Pattern Recognition}, pages 9707--9716, 2023{\natexlab{c}}.

\bibitem[Garg et~al.(2025)Garg, Kumar, and Rawat]{garg2025stpro}
Aaryan Garg, Akash Kumar, and Yogesh~S Rawat.
\newblock Stpro: Spatial and temporal progressive learning for weakly supervised spatio-temporal grounding.
\newblock In \emph{Proceedings of the Computer Vision and Pattern Recognition Conference}, pages 3384--3394, 2025.

\bibitem[He et~al.(2015)He, Zhang, Ren, and Sun]{resnet}
Kaiming He, X. Zhang, Shaoqing Ren, and Jian Sun.
\newblock Deep residual learning for image recognition.
\newblock \emph{2016 IEEE Conference on Computer Vision and Pattern Recognition (CVPR)}, pages 770--778, 2015.

\bibitem[Ho et~al.(2020)Ho, Jain, and Abbeel]{ddpm}
Jonathan Ho, Ajay Jain, and Pieter Abbeel.
\newblock Denoising diffusion probabilistic models.
\newblock \emph{Advances in neural information processing systems}, 33:\penalty0 6840--6851, 2020.

\bibitem[Ho et~al.(2022)Ho, Salimans, Gritsenko, Chan, Norouzi, and Fleet]{Ho2022VideoDM}
Jonathan Ho, Tim Salimans, Alexey Gritsenko, William Chan, Mohammad Norouzi, and David~J. Fleet.
\newblock Video diffusion models.
\newblock \emph{ArXiv}, abs/2204.03458, 2022.

\bibitem[Huang et~al.(2021)Huang, Xue, Shen, Tian, Li, Huang, and Hua]{huang20213d}
Zhen Huang, Dixiu Xue, Xu Shen, Xinmei Tian, Houqiang Li, Jianqiang Huang, and Xian-Sheng Hua.
\newblock 3d local convolutional neural networks for gait recognition.
\newblock In \emph{2021 IEEE/CVF International Conference on Computer Vision (ICCV)}, pages 14900--14909, 2021.

\bibitem[Jain et~al.(1997)Jain, Hong, and Bolle]{jain1997line}
Anil Jain, Lin Hong, and Ruud Bolle.
\newblock On-line fingerprint verification.
\newblock \emph{IEEE transactions on pattern analysis and machine intelligence}, 19\penalty0 (4):\penalty0 302--314, 1997.

\bibitem[Kumar and Rawat(2022)]{Kumar_2022_CVPR}
Akash Kumar and Yogesh~Singh Rawat.
\newblock End-to-end semi-supervised learning for video action detection.
\newblock In \emph{Proceedings of the IEEE/CVF Conference on Computer Vision and Pattern Recognition (CVPR)}, 2022.

\bibitem[Kumar et~al.(2023)Kumar, Kumar, Vineet, and Rawat]{kumar2023benchmarking}
Akash Kumar, Ashlesha Kumar, Vibhav Vineet, and Yogesh~Singh Rawat.
\newblock Benchmarking self-supervised video representation learning.
\newblock \emph{Neural Information Processing Systems 4th Workshop on Self-Supervised Learning: Theory and Practice}, 2023.

\bibitem[Kumar et~al.(2025{\natexlab{a}})Kumar, Kira, and Rawat]{kumar2025contextual}
Akash Kumar, Zsolt Kira, and Yogesh~Singh Rawat.
\newblock Contextual self-paced learning for weakly supervised spatio-temporal video grounding.
\newblock \emph{Proceedings of the International Conference on Learning Representations (ICLR)}, 2025{\natexlab{a}}.

\bibitem[Kumar et~al.(2025{\natexlab{b}})Kumar, Kumar, Vineet, and Rawat]{Kumar_2025_CVPR}
Akash Kumar, Ashlesha Kumar, Vibhav Vineet, and Yogesh~S Rawat.
\newblock A large-scale analysis on contextual self-supervised video representation learning.
\newblock In \emph{Proceedings of the IEEE/CVF Conference on Computer Vision and Pattern Recognition (CVPR) Workshops}, pages 670--681, 2025{\natexlab{b}}.

\bibitem[Kumar et~al.(2025{\natexlab{c}})Kumar, Mitra, and Rawat]{kumar2025stable}
Akash Kumar, Sirshapan Mitra, and Yogesh~Singh Rawat.
\newblock Stable mean teacher for semi-supervised video action detection.
\newblock In \emph{Proceedings of the AAAI Conference on Artificial Intelligence}, pages 4419--4427, 2025{\natexlab{c}}.

\bibitem[Li et~al.(2018)Li, Zhang, Chen, and Feng]{li2018facial}
Defang Li, Min Zhang, Weifu Chen, and Guocan Feng.
\newblock Facial attribute editing by latent space adversarial variational autoencoders.
\newblock In \emph{2018 24th International Conference on Pattern Recognition (ICPR)}, pages 1337--1342. IEEE, 2018.

\bibitem[Li et~al.(2024)Li, Xu, Wu, Xiong, Deng, Ji, Huang, Feng, Ding, and Hooi]{li2024id}
Shen Li, Jianqing Xu, Jiaying Wu, Miao Xiong, Ailin Deng, Jiazhen Ji, Yuge Huang, Wenjie Feng, Shouhong Ding, and Bryan Hooi.
\newblock Id3: Identity-preserving-yet-diversified diffusion models for synthetic face recognition.
\newblock \emph{arXiv preprint arXiv:2409.17576}, 2024.

\bibitem[Liang and Rawat(2025)]{Liang_2025_CVPR}
Xin Liang and Yogesh~S Rawat.
\newblock Differ: Disentangling identity features via semantic cues for clothes-changing person re-id.
\newblock In \emph{Proceedings of the Computer Vision and Pattern Recognition Conference (CVPR)}, pages 13980--13989, 2025.

\bibitem[Lin et~al.(2021)Lin, Zhang, and Yu]{gaitgl}
Beibei Lin, Shunli Zhang, and Xin Yu.
\newblock Gait recognition via effective global-local feature representation and local temporal aggregation.
\newblock In \emph{Proceedings of the IEEE/CVF International Conference on Computer Vision}, pages 14648--14656, 2021.

\bibitem[Ma et~al.(2023{\natexlab{a}})Ma, Ye, Fan, and Yu]{ma2023pedestrian}
Jingzhe Ma, Dingqiang Ye, Chao Fan, and Shiqi Yu.
\newblock Pedestrian attribute editing for gait recognition and anonymization.
\newblock \emph{arXiv preprint arXiv:2303.05076}, 2023{\natexlab{a}}.

\bibitem[Ma et~al.(2023{\natexlab{b}})Ma, Fu, Zheng, Cao, Hu, and Huang]{Ma_2023_CVPR}
Kang Ma, Ying Fu, Dezhi Zheng, Chunshui Cao, Xuecai Hu, and Yongzhen Huang.
\newblock Dynamic aggregated network for gait recognition.
\newblock In \emph{Proceedings of the IEEE/CVF Conference on Computer Vision and Pattern Recognition (CVPR)}, pages 22076--22085, 2023{\natexlab{b}}.

\bibitem[Modi et~al.(2022)Modi, Rana, Kumar, Tirupattur, Vyas, Rawat, and Shah]{modi2022video}
Rajat Modi, Aayush~Jung Rana, Akash Kumar, Praveen Tirupattur, Shruti Vyas, Yogesh~Singh Rawat, and Mubarak Shah.
\newblock Video action detection: Analysing limitations and challenges.
\newblock \emph{2022 IEEE/CVF Conference on Computer Vision and Pattern Recognition Workshops (CVPRW)}, pages 4907--4916, 2022.

\bibitem[Nixon and Carter(2006)]{nixon2006automatic}
Mark~S Nixon and John~N Carter.
\newblock Automatic recognition by gait.
\newblock \emph{Proceedings of the IEEE}, 94\penalty0 (11):\penalty0 2013--2024, 2006.

\bibitem[Pathak and Rawat(2025{\natexlab{a}})]{Pathak_2025_ICCV}
Priyank Pathak and Yogesh~S Rawat.
\newblock Colors see colors ignore: Clothes changing reid with color disentanglement.
\newblock In \emph{Proceedings of the IEEE/CVF International Conference on Computer Vision (ICCV)}, 2025{\natexlab{a}}.

\bibitem[Pathak and Rawat(2025{\natexlab{b}})]{pathak2025coarse}
Priyank Pathak and Yogesh~S Rawat.
\newblock Coarse attribute prediction with task agnostic distillation for real world clothes changing reid.
\newblock In \emph{36th British Machine Vision Conference 2025, {BMVC} 2025, Sheffield, UK, November 24-27, 2025}. {BMVA} Press, 2025{\natexlab{b}}.

\bibitem[Shahreza and Marcel(2024)]{shahreza2024hyperface}
Hatef~Otroshi Shahreza and Sebastien Marcel.
\newblock Hyperface: Generating synthetic face recognition datasets by exploring face embedding hypersphere.
\newblock \emph{arXiv preprint arXiv:2411.08470}, 2024.

\bibitem[Singh et~al.(2024)Singh, Rana, Kumar, Vyas, and Rawat]{Singh_Rana_Kumar_Vyas_Rawat_2024}
Ayush Singh, Aayush~J Rana, Akash Kumar, Shruti Vyas, and Yogesh~Singh Rawat.
\newblock Semi-supervised active learning for video action detection.
\newblock \emph{Proceedings of the AAAI Conference on Artificial Intelligence}, 38\penalty0 (5):\penalty0 4891--4899, 2024.

\bibitem[Stypulkowski et~al.(2024)Stypulkowski, Vougioukas, He, Zieba, Petridis, and Pantic]{stypulkowski2024diffused}
Michal Stypulkowski, Konstantinos Vougioukas, Sen He, Maciej Zieba, Stavros Petridis, and Maja Pantic.
\newblock Diffused heads: Diffusion models beat gans on talking-face generation.
\newblock In \emph{Proceedings of the IEEE/CVF Winter Conference on Applications of Computer Vision}, pages 5091--5100, 2024.

\bibitem[Takemura et~al.(2018)Takemura, Makihara, Muramatsu, Echigo, and Yagi]{oumvlp}
Noriko Takemura, Yasushi Makihara, Daigo Muramatsu, Tomio Echigo, and Yasushi Yagi.
\newblock Multi-view large population gait dataset and its performance evaluation for cross-view gait recognition.
\newblock \emph{IPSJ transactions on Computer Vision and Applications}, 10:\penalty0 1--14, 2018.

\bibitem[Tu and Chen(2019)]{tu2019facial}
Ching-Ting Tu and Yi-Fu Chen.
\newblock Facial image inpainting with variational autoencoder.
\newblock In \emph{2019 2nd international conference of intelligent robotic and control engineering (IRCE)}, pages 119--122. IEEE, 2019.

\bibitem[Wang et~al.(2023{\natexlab{a}})Wang, Liu, Liang, and Wang]{Wang2023HierarchicalSR}
Lei Wang, Bo Liu, Fangfang Liang, and Bin Wang.
\newblock Hierarchical spatio-temporal representation learning for gait recognition.
\newblock \emph{ArXiv}, abs/2307.09856, 2023{\natexlab{a}}.

\bibitem[Wang et~al.(2023{\natexlab{b}})Wang, Guo, Lin, Yang, Zhu, Li, Zhang, and Yu]{Wang2023DyGaitED}
Ming-Zhen Wang, Xianda Guo, Beibei Lin, Tian Yang, Zhenguo Zhu, Lincheng Li, Shunli Zhang, and Xin Yu.
\newblock Dygait: Exploiting dynamic representations for high-performance gait recognition.
\newblock \emph{ArXiv}, abs/2303.14953, 2023{\natexlab{b}}.

\bibitem[Yang et~al.(2023)Yang, Jiang, Liu, and Loy]{yang2023styleganex}
Shuai Yang, Liming Jiang, Ziwei Liu, and Chen~Change Loy.
\newblock Styleganex: Stylegan-based manipulation beyond cropped aligned faces.
\newblock In \emph{Proceedings of the IEEE/CVF International Conference on Computer Vision}, pages 21000--21010, 2023.

\bibitem[Yi et~al.(2013)Yi, Lei, and Li]{yi2013towards}
Dong Yi, Zhen Lei, and Stan~Z Li.
\newblock Towards pose robust face recognition.
\newblock In \emph{Proceedings of the IEEE conference on computer vision and pattern recognition}, pages 3539--3545, 2013.

\bibitem[Yu et~al.(2006)Yu, Tan, and Tan]{casia}
Shiqi Yu, Daoliang Tan, and Tieniu Tan.
\newblock A framework for evaluating the effect of view angle, clothing and carrying condition on gait recognition.
\newblock In \emph{18th international conference on pattern recognition (ICPR'06)}, pages 441--444. IEEE, 2006.

\bibitem[Yu et~al.(2017)Yu, Chen, Garcia~Reyes, and Poh]{yu2017gaitgan}
Shiqi Yu, Haifeng Chen, Edel~B Garcia~Reyes, and Norman Poh.
\newblock Gaitgan: Invariant gait feature extraction using generative adversarial networks.
\newblock In \emph{Proceedings of the IEEE conference on computer vision and pattern recognition workshops}, pages 30--37, 2017.

\bibitem[Yu et~al.(2019)Yu, Liao, An, Chen, Reyes, Huang, and Poh]{yu2017gaitgan2}
Shiqi Yu, Rijun Liao, Weizhi An, Haifeng Chen, Edel B.~García Reyes, Yongzhen Huang, and Norman Poh.
\newblock Gaitganv2: Invariant gait feature extraction using generative adversarial networks.
\newblock \emph{Pattern Recognit.}, 87:\penalty0 179--189, 2019.

\bibitem[Zhang et~al.(2023)Zhang, Wu, Liu, Zhao, Ran, Gu, Gao, and Shou]{zhang2023show}
David~Junhao Zhang, Jay~Zhangjie Wu, Jia-Wei Liu, Rui Zhao, Lingmin Ran, Yuchao Gu, Difei Gao, and Mike~Zheng Shou.
\newblock Show-1: Marrying pixel and latent diffusion models for text-to-video generation.
\newblock \emph{arXiv preprint arXiv:2309.15818}, 2023.

\bibitem[Zhou et~al.(2022)Zhou, Wang, Yan, Lv, Zhu, and Feng]{zhou2022magicvideo}
Daquan Zhou, Weimin Wang, Hanshu Yan, Weiwei Lv, Yizhe Zhu, and Jiashi Feng.
\newblock Magicvideo: Efficient video generation with latent diffusion models.
\newblock \emph{arXiv preprint arXiv:2211.11018}, 2022.

\bibitem[Zhu et~al.(2021)Zhu, Guo, Yang, Huang, Deng, Huang, Du, Lu, and Zhou]{grew}
Zheng Zhu, Xianda Guo, Tian Yang, Junjie Huang, Jiankang Deng, Guan Huang, Dalong Du, Jiwen Lu, and Jie Zhou.
\newblock Gait recognition in the wild: A benchmark.
\newblock In \emph{Proceedings of the IEEE/CVF international conference on computer vision}, pages 14789--14799, 2021.

\end{thebibliography}
